# Local Directional Gradient Pattern: A Local Descriptor for Face Recognition

Soumendu Chakraborty, *Student Member, IEEE*, Satish Kumar Singh, *Senior Member, IEEE* and Pavan Chakraborty, *Member, IEEE*

*Abstract*—In this paper a local pattern descriptor in high order derivative space is proposed for face recognition. The proposed local directional gradient pattern (LDGP) is a 1D local micropattern computed by encoding the relationships between the higher order derivatives of the reference pixel in four distinct directions. The proposed descriptor identifies the relationship between the high order derivatives of the referenced pixel in four different directions to compute the micropattern which corresponds to the local feature. Proposed descriptor considerably reduces the length of the micropattern which consequently reduces the extraction time and matching time while maintaining the recognition rate. Results of the extensive experiments conducted on benchmark databases AT&T, Extended Yale B and CMU-PIE show that the proposed descriptor significantly reduces the extraction as well as matching time while the recognition rate is almost similar to the existing state of the art methods.

## I. INTRODUCTION

FACE image analysis has become an active area of research for past 15 years. Face recognition, facial image retrieval and statistical feature extraction are some of the recurring problems that have been addressed in the recent literature. A descriptor is a statistical feature that uniquely identifies an image. Eigen-face[1], Fischer-face[3], Principle Component Analysis (PCA)[18] and Linear Discriminant Analysis (LDA)[2], are some of the holistic approaches extensively studied in the literature. Defining a robust descriptor which effectively eliminates variations in poses and illumination is a big challenge. Local pattern descriptors [4]-[6] are gaining more attention in recent years. One of the most commonly used local descriptors in facial image recognition is local binary pattern (LBP) [7]-[10]. LBP is the first order local pattern descriptor which encodes the relationship between the locally identified reference pixel and the neighboring pixels. A binary micropattern is generated by the encoding function which represents the local microstructure in the image. These micropatterns modeled using spatial histogram to represent the local feature of the image. More discriminating features are extracted in higher order derivative space which improves the recognition accuracy [11]. Local derivative pattern (LDP) [11] is a local pattern descriptor which generates more discriminating information in the higher order derivative space and achieves better results as compared to LBP. Most recently a local pattern descriptor called local vector pattern (LVP) has been proposed which identifies the relationships amongst the pixels in local neighborhood of the higher order derivative space [12]. In real time face recognition for huge databases the problem of feature length increasing associated with LDP as well as LVP degrades the performance of the recognition system. Increasing feature length of a descriptor increases the time to extract and match the feature of all the images for all the classes in a database. The problem of increase in extraction time and matching time worsens with the increasing size of database and image as well.

In this paper a higher order local pattern descriptor called local directional gradient pattern (LDGP) is proposed which identifies relationship amongst reference pixel and four neighbors in four directions and generates micropatterns in the higher order derivative space. The proposed descriptor effectively reduces the feature length and achieves lower feature extraction as well as matching time. The reduced feature length does not affect recognition accuracy and achieves comparable recognition rates.

The remaining sections of this paper are organized as follows. Section II briefly introduces existing higher order local pattern descriptors and elaborates the effect of increasing size of images and growing size of database on extraction and matching time. Proposed descriptor and its higher order definitions are elaborated in Section III. Section IV elaborates the similarity measure used in the experiments to analyze the accuracies of the proposed and the existing higher order descriptors. Experimental framework is explained in section V. Section V also illustrates the experimental results. We conclude in section VI.

## II. HIGHER ORDER LOCAL PATTERN DESCRIPTOR

Higher order local descriptors extract more discriminative information and perform well with variations in pose as well as illumination [11]. In this section a brief overview of two higher order descriptors are given.

### A. Local Derivative Pattern

LDP is an expansion of LBP in higher order derivative space. LDP Defines $(n-1)^{th}$ order derivatives in four different directions for a reference pixel. $n^{th}$ order LDP is calculated from $(n-1)^{th}$ order derivatives using an encoding function in four derivative directions. $n^{th}$ order LDP is a binary string generated by concatenating four binary strings of 8 bits each corresponding to micropatterns in $(n-1)^{th}$ derivative space in different directions. Hence the length of LDP for a reference pixel is 32 bit which is large enough to increase the extraction time and matching time considerably for a large dataset.

*B. Local Vector Pattern*

LVP is an improvement over local tetra pattern (LTrP) [13] which identifies the relationship of the reference pixel with 8 neighbors at a distance D and encodes them using comparative space transformation (CST). LVP performs well in lighting, pose, and illumination variations [12]. Fundamental contribution of LVP is the CST which encodes the pair wise derivatives in four distinct directions [12]. The feature length for LVP is also 32 bits, which increases the extraction as well as matching time for face recognition.

LDP and LVP are the $n^{th}$ order local descriptor which encodes the $(n-1)^{th}$ order derivatives for all eight neighbors of reference pixel in $0°, 45°, 90°, and\ 135°$ directions. LDP and LVP perform better than LBP as more discriminative features are encoded in higher order derivative space. Length of the local feature encoded using LDP as well as LVP descriptors are four times the length of the local feature encoded using LBP or LDGP which affects the extraction as well as matching time of face recognition system.

Let $t_e$ be the feature extraction time which is directly proportional to the size of the image $S = M \times N$, total number of the images in the database $\Gamma = \sum_{i=1}^{T} w_i$ where $T$ is the number of classes in the database with $w_i$ number of images in the $i^{th}$ class, and final feature length $\hat{l}$. Hence

$$t_e = K_1 \times S \times \Gamma \times \hat{l} \qquad (1)$$

where $K_1$ is the proportionality constant.

Similarly matching time $t_m$ is directly proportional to the size of the image $S$, total number of the images in the database $\Gamma$, and final feature length $\hat{l}$. Hence

$$t_m = K_2 \times S \times \Gamma \times \hat{l} \qquad (2)$$

where $K_2$ is the proportionality constant.

From equation (1) and (2) it is evident that extraction time and match time have a nonlinear relationship with feature length. Thus feature length increasing adversely affects the recognition system with respect to extraction time and match time.

Extraction times of LDP and LVP shown in Fig.4(a) and Fig.4(c) illustrate the problem of increasing extraction time with increasing size of the database as well as the size of images. Feature length increasing not only affects the extraction time it also affects the match time of a descriptor used in a recognition system. Fig.4(b) and Fig.4(d) illustrate the effect of increased feature lengths of LDP and LVP on match time of the framework with increasing size of images and database respectively.

### III. Proposed Local directional gradient pattern

Proposed LDGP is a local descriptor which encodes $(n-1)^{th}$ order derivatives for four neighbors of reference pixel in $0°, 45°, 90°, and\ 135°$ directions. New encoding function of LDGP effectively code the relationship of neighboring pixels in higher order derivative space without losing the discriminative information in higher order derivative space and significantly reduces the length of the local feature which is less than the feature length computed using LBP, LVP and LDP.

Four first order directional derivatives of the image $I(\mathbb{R})$ in $0°, 45°, 90°, and\ 135°$ directions for an arbitrary reference point $\mathbb{R}_0$ are computed as computed in [11] and [12] as follows

$$I'_{0°}(\mathbb{R}_0) = I(\mathbb{R}_0) - I(\mathbb{R}_1) \qquad (3)$$

$$I'_{45°}(\mathbb{R}_0) = I(\mathbb{R}_0) - I(\mathbb{R}_2) \qquad (4)$$

$$I'_{90°}(\mathbb{R}_0) = I(\mathbb{R}_0) - I(\mathbb{R}_3) \qquad (5)$$

$$I'_{135°}(\mathbb{R}_0) = I(\mathbb{R}_0) - I(\mathbb{R}_4) \qquad (6)$$

Second order $LDGP^2(.)$ is encoded with the encoding function $C(.,.)$ as

$$LDGP^2(\mathbb{R}_0) = \left\{ C\left(I'_{0°}(\mathbb{R}_0), I'_{45°}(\mathbb{R}_0)\right), C\left(I'_{0°}(\mathbb{R}_0), I'_{90°}(\mathbb{R}_0)\right), \\ C\left(I'_{0°}(\mathbb{R}_0), I'_{135°}(\mathbb{R}_0)\right), C\left(I'_{45°}(\mathbb{R}_0), I'_{90°}(\mathbb{R}_0)\right), \\ C\left(I'_{45°}(\mathbb{R}_0), I'_{135°}(\mathbb{R}_0)\right), C\left(I'_{90°}(\mathbb{R}_0), I'_{135°}(\mathbb{R}_0)\right) \right\} \qquad (7)$$

Encoding function $C(.,.)$ codes the pair of derivative directions of the reference pixel as

$$C\left(I'_\theta(\mathbb{R}_0), I'_\alpha(\mathbb{R}_0)\right) = \begin{cases} 1, & if\ I'_\theta(\mathbb{R}_0) > I'_\alpha(\mathbb{R}_0) \\ 0, & else \end{cases} \qquad (8)$$

It is clear from equation (7) that LDGP computes the binary pattern of 6 bits for each pixel points in the image from the corresponding derivatives in $0°, 45°, 90°, and\ 135°$ directions.

Third order LDGP is computed from second order derivatives. Second order derivatives are calculated from first order derivatives as

$$I''_{0°}(\mathbb{R}_0) = I'_{0°}(\mathbb{R}_0) - I'_{0°}(\mathbb{R}_1) \qquad (9)$$

$$I''_{45°}(\mathbb{R}_0) = I'_{45°}(\mathbb{R}_0) - I'_{45°}(\mathbb{R}_2) \qquad (10)$$

$$I''_{90°}(\mathbb{R}_0) = I'_{90°}(\mathbb{R}_0) - I'_{90°}(\mathbb{R}_3) \qquad (11)$$

$$I''_{135°}(\mathbb{R}_0) = I'_{135°}(\mathbb{R}_0) - I'_{135°}(\mathbb{R}_4) \qquad (12)$$

Third order $LDGP^3(.)$ is encoded with the encoding function $C(.,.)$ as

$$LDGP^3(\mathbb{R}_0) = \{C(I_{0°}''(\mathbb{R}_0), I_{45°}''(\mathbb{R}_0)), C(I_{0°}''(\mathbb{R}_0), I_{90°}''(\mathbb{R}_0)),$$
$$C(I_{0°}''(\mathbb{R}_0), I_{135°}''(\mathbb{R}_0)), C(I_{45°}''(\mathbb{R}_0), I_{90°}''(\mathbb{R}_0)),$$
$$C(I_{45°}''(\mathbb{R}_0), I_{135°}''(\mathbb{R}_0)), C(I_{90°}''(\mathbb{R}_0), I_{135°}''(\mathbb{R}_0))\} \quad (13)$$

Similarly $n^{th}$ order $LDGP^n(.)$ is computed from $(n-1)^{th}$ order derivative as

$$LDGP^n(\mathbb{R}_0) =$$
$$\{C(I_{0°}^{n-1}(\mathbb{R}_0), I_{45°}^{n-1}(\mathbb{R}_0)), C(I_{0°}^{n-1}(\mathbb{R}_0), I_{90°}^{n-1}(\mathbb{R}_0)),$$
$$C(I_{0°}^{n-1}(\mathbb{R}_0), I_{135°}^{n-1}(\mathbb{R}_0)), C(I_{45°}^{n-1}(\mathbb{R}_0), I_{90°}^{n-1}(\mathbb{R}_0)),$$
$$C(I_{45°}^{n-1}(\mathbb{R}_0), I_{135°}^{n-1}(\mathbb{R}_0)), C(I_{90°}^{n-1}(\mathbb{R}_0), I_{135°}^{n-1}(\mathbb{R}_0))\} \quad (14)$$

Encoding function $C(.,.)$ codes the pair of $(n-1)^{th}$ derivative directions of the reference pixel as

$$C\left(I_\theta^{n-1}(\mathbb{R}_0), I_\alpha^{n-1}(\mathbb{R}_0)\right)$$
$$= \begin{cases} 1, & if\ I_\theta^{n-1}(\mathbb{R}_0) > I_\alpha^{n-1}(\mathbb{R}_0) \\ 0, & else \end{cases} \quad (15)$$

$LDGP^n(\mathbb{R}_0)$ is the $n^{th}$ order LDGP obtained from $(n-1)^{th}$ order derivatives. More discriminating features are extracted from the higher order local pattern to explore the detailed texture information. However local patterns are susceptible to noise with increasing order $n$ [12]. Hence we compare the performance of LDP and LVP with proposed LDGP in the $1^{st}$ order derivative space.

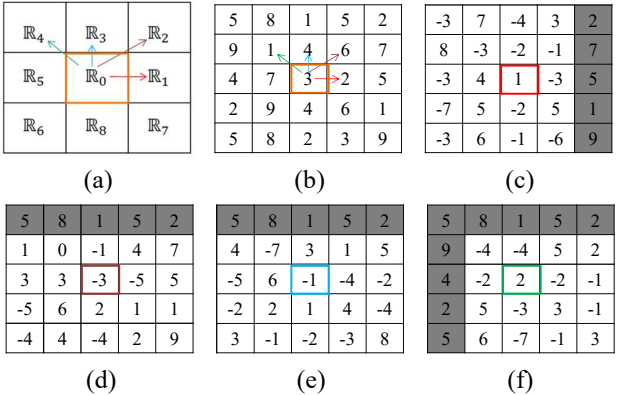

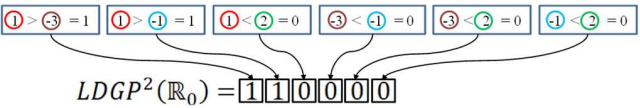

Fig. 1. Example illustrating the computation of first order derivatives in $0°, 45°, 90° and\ 135°$ directions.

Fig. 2. Calculation of second order $LDGP$ for the reference point $\mathbb{R}_0$ in the sample image from first order derivatives in $0°, 45°, 90° and\ 135°$ directions.

Reference pixel (point) and four neighbors in four directions in a sample image are shown in Fig.1.(a). Value of the reference pixel is 3 as shown by the orange box in Fig.1.(b). Neighboring pixels in different directions are shown by the different colored arrows. Neighboring pixels in $0°$, $45°, 90°$, and $135°$ are 2, 6, 4 and 1 respectively for the reference pixel 3. Fig.1(c) to Fig.1(f) show the $1^{st}$ order derivatives in four distinct directions computed by subtracting the neighboring pixel in respective directions from the reference pixel. $1^{st}$ order derivative of reference pixel 3 in $0°$ direction is $3 - 2 = 1$ as shown in Fig 1.(c) within red box. Similarly $1^{st}$ order derivatives of reference pixel 3 in $45°, 90°$, and $135°$ directions are $-3, -1$ and 2 shown in Fig.1(d), Fig.1(e) and Fig.1(f) respectively. Pixels which do not change in the respective derivatives are shown by the shaded portion of the derivative image. Fig 2, illustrates the use of coding scheme in calculation of the $2^{nd}$ order LDGP for the reference pixel. Relationship between $1^{st}$ order derivative in $0°$ direction and $45°$ direction is encoded as 1 using equation (8). Similarly relationship between $1^{st}$ order derivative in $0°$ direction and $135°$ direction is encoded as 0. There are four $1^{st}$ order derivative points for a particular reference pixel and six possible combinations of two derivative points each. Finally encoding all possible combinations the $2^{nd}$ order LDGP for the reference pixel 3 is generated as 110000 as shown in Fig 2. Visualization of the higher order descriptors are shown in Fig.3. Fig.3 (d-f) shows that higher order LDGP is able to extract more discriminative information from original grayscale image shown in Fig.3(a). $2^{nd}$ order LDP and LVP in $0°$ direction are shown in Fig.3(b) and Fig.3(c) respectively.

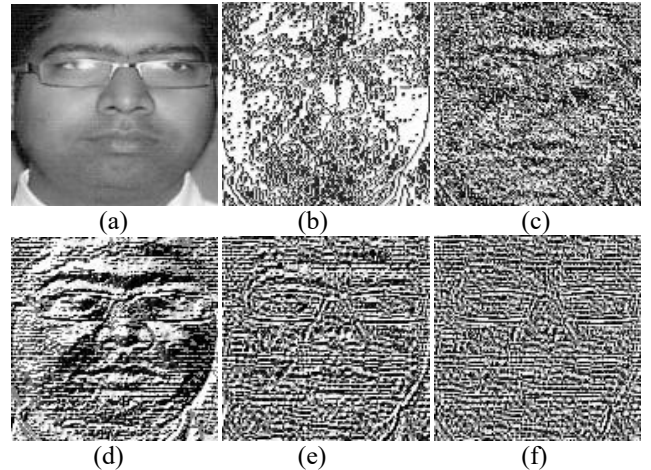

Fig. 3. Higher order LDGP, LVP and LDP (a) original grayscale image, (b) $2^{nd}$ order LDP in $0°$ direction, (c) $2^{nd}$ order LVP in $0°$ direction, (d) $2^{nd}$ order LDGP, (e) $3^{rd}$ order LDGP, (f) $4^{th}$ order LDGP

## IV. HISTOGRAM AND SIMILARITY MEASURE

Local descriptors are modeled using the spatial histogram in face recognition as it is robust in pose and illumination variations [10]. Hence spatial histogram is used to model the higher order LDGP values. Higher order LDGP is calculated from the lower order derivatives as binary strings of 6 bits. These strings are then converted into the equivalent decimal values. Resulting LDGP matrix is divided into U local regions. These regions are represented as $V_1, V_2, V_3 \dots V_U$. Spatial histogram is computed on these local regions and concatenated to represent the histogram feature of the facial image. Spatial histogram $HLDGP(.)$ on local region is computed as

$$HLDGP(i) = \{H_{LDGP}(V_i)|i = 1, 2, \dots U\} \quad (16)$$

where $H_{LDGP}(V_i)$ represents the spatial histogram of the $i^{th}$ local region.

Several similarity measures [10] are available to match the spatial histograms. In this paper, taxicab distance ($L1\ distance$) is used to measure the similarity of two histograms as it performs better than other measures on the datasets used in the experiments. Similarity measure $S_{L1}(.,.)$ is defined as

$$S_{L1}(X,Y) = \sum_{i=0}^{q} |x_i - y_i| \quad (17)$$

where $S_{L1}(X,Y)$ is the $L1$ distance computed on two vectors $X = (x_1, \dots, x_q)$ and $Y = (y_1, \dots, y_q)$. Nearest one neighbor (1 NN) classifier is used as used in [11] to compute the minimum $L1$ distance between the probe image and the gallery images. As similar regions of the probe and gallery images are effectively identified by 1NN classifier with optimal computational cost [11].

## V. Experimental Results

Experiments are conducted on three publicly available datasets AT&T [14], Extended Yale B [15] [16] and CMU-PIE [17]. The main objective of the experiments are to show that the proposed descriptor effectively reduces the feature length to significantly reduce the extraction and more importantly match time without affecting the accuracy of the recognition system. To compare the performance of the proposed method different environmental parameters are considered such as pose, illumination, different lighting conditions and expressions. Images are partitioned into sub-regions of varying size to compute the histogram of 8, 16 and 32 bins. Uniform quantization is used to reduce the size of the histogram to 8, 16 and 32 bins.

### A. Performance analysis on AT&T database

AT&T database contains 10 different images of 40 subjects in different pose. For some the subjects images are taken with varying lighting, different expressions (open or closed eyes, smiling or non-smiling), and facial details (glasses or no-glasses). A dark homogeneous background is used for all 40 subjects.

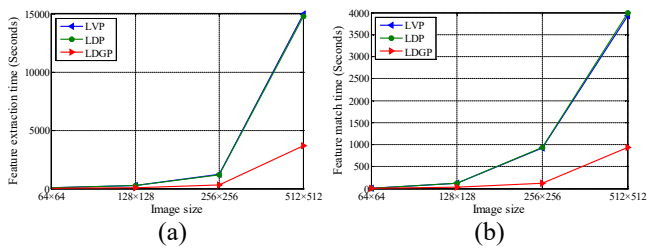

(a)                    (b)

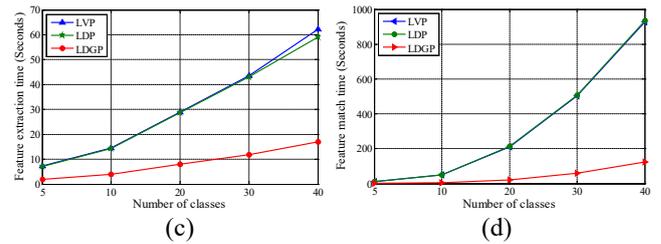

(c)                    (d)

Fig. 4. Comparative computation time of LDGP, LVP and LDP on AT&T face database (a) extraction time with varying size of images, (b) match time with varying size of images, (c) extraction time with varying size of database and fixed image size of $64 \times 64$, (d) match time with varying size of database and fixed image size of $256 \times 256$.

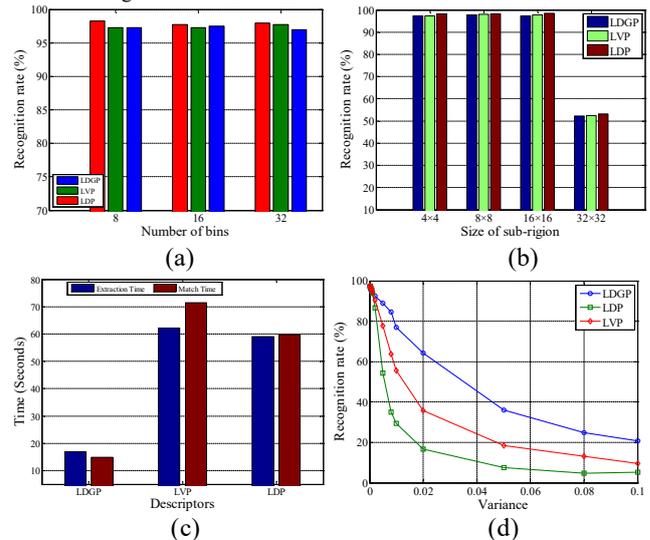

(a)                    (b)

(c)                    (d)

Fig. 5. Comparative recognition rates of LDGP, LVP and LDP on AT&T face datasets with (a) different number of histogram bins, (b) different size of sub-regions. (c) Extraction time and Match time of LDGP, LVP and LDP. (d) Comparative recognition rates of LDGP, LVP and LDP with different values of variance

Experiments are conducted using each image as probe and remaining images as gallery for all the images in the dataset. 1st nearest neighbor is computed by comparing the distances between probe and the gallery images calculated using equation (17). If the class of the nearest neighbor from the gallery is same as the class of the probe image then it is considered as a match. Recognition rate for the experimental framework is calculated as

$$\gamma = \frac{N_m}{N_t} \times 100 \quad (18)$$

where $N_m$ is the total number of matches and $N_t$ is the total number of images in the database.

To illustrate the performance deficiencies of LDP and VVP with respect to time experiments are conducted on images with different size and with varying size of database. Extraction time and match time are calculated for varying size of images of 40 subjects in AT&T database. Match time is the time taken to match each image taken as probe and remaining images as gallery. Hence match time in real sense is the time taken to compute $400 \times 399$ distances and to perform $400 \times 399$ comparisons. Fig.4(a) shows the feature extraction time for 10 images each of varying size for 40 subjects and Fig.4(b) shows the corresponding match time. Extraction times of LDGP, LVP and LDP for images of size $512 \times 512$ are 3665.87, 14988.23 and 14802.57 seconds respectively. Hence it is evident that the extraction time of LVP and LDP are approximately 4 times the extraction time of LDGP. Match times of LDGP, LVP and LDP for images of size $512 \times 512$ are 939.69, 3936.17 and 3990.81 seconds respectively. LDGP again outperforms LVP and LDP with respect to match time. To illustrate the effect of feature length increasing on growing size of database the extraction time and match time are calculated with images of fixed size and varying size of database. Fig.4(c) and Fig.4(d) show the extraction time and match time with varying number of classes (subjects). Extraction times of LDGP, LVP and LDP for images of size $64 \times 64$ of 40 classes (subjects) are 16.97, 62.24 and 59.08 seconds respectively whereas match times of LDGP, LVP and LDP for images of size $256 \times 256$ of 40 classes are 124.44, 926.25 and 934.20 seconds respectively. Fig.4 shows how severe is the effect of large feature length of LDP and LVP on performance of the recognition system and how LDGP overcomes this deficiency.

Extraction time shown in Fig.5(c) is the time taken to compute the features of all the images in the dataset using $4 \times 4$ sub-regions and histogram of 8 bins and match time shown in Fig.5(c) is the time taken to calculate the recognition rates with $4 \times 4$ sub-regions and histogram of 8 bins. Extraction times of LDGP, LVP and LDP are 16.97, 62.24 and 59.08 seconds respectively and match times of LDGP, LVP and LDP are 14.78, 71.36 and 59.72 seconds respectively. Extraction time and match time of LDP as well as LVP are approximately 4 times the extraction time and match time of LDGP.

Recognition rates are calculated using features of probe as well as gallery images with reduced histograms of 8, 16 and 32 bins where each image is partitioned into $4 \times 4$ sub-regions. Fig.5(a) shows the recognition rates of different descriptors for different number of bins. Recognition rates of LDGP, LVP and LDP are 97.50%, 97.25% and 97.75% respectively with features computed using local histogram of 16 bins. The difference in recognition rates of LDGP and LDP is negligible as compared to the difference in extraction and match time of LDGP and LDP as shown in Fig.5(c). Recognition rate of LDGP is better than LVP for local histogram of 16 bins.

Experiments are conducted on images with varying sub-regions of size $4 \times 4$, $8 \times 8$, $16 \times 16$ and $32 \times 32$. To compute the feature fixed size local histograms of 8 bins are used. Fig.5(b) shows the recognition rates of different descriptors for varying size of sub-regions. Recognition rates of LDGP, LVP and LDP are 97.75%, 98% and 98.25% respectively with features computed using $8 \times 8$ sub-regions. The difference in recognition rates of LDGP, LDP and LDGP, LVP are 0.50% and 0.25% respectively which are comparable when we consider the 4 times reduction in extraction time and match time achieved by the proposed LDGP as shown in Fig.5(c). It is evident from Fig.5(a) and Fig.5(b) that almost similar recognition rates are achieved by all three descriptors for different sub-region size and different number of bins. Hence remaining experiments are conducted with sub-region size of $4 \times 4$ and reduced histogram of 8 bins.

Features with zero mean and different values of variance ($\sigma^2$) of Gaussian white noise are extracted with LDGP, LVP and LDP descriptors in the first order derivative space. Recognition rates are computed using these features as probe and the features of the original images as gallery. Recognition rates for images with different values of variance are shown in Fig. 5(d). Fig. 5(d) illustrates that the proposed LDGP performs better than LDP and LVP in the presence of higher levels of noise. Fig.5(d) clearly shows that with 99% increase in variance of the noise the reduction in recognition accuracies of LDGP, LVP and LDP are 77.92%, 89.92% and 94.41% respectively. Hence we conclude that the reduction in accuracy of LDGP is less compared to LVP and LDP for higher values of variance.

## B. Performance analysis on Extended Yale B database

Extended Yale B is used in the experiments as it contains images with 64 different illumination variations of 38 subjects. It is a benchmark dataset used to test the robustness of a descriptor against illumination variations. Cropped version of the Extended Yale B dataset is used in the experiments.

Extraction time and match time over 2432 images for three descriptors are shown in Fig.6(a). Extraction time is computed for $4 \times 4$ sub-regions and histogram of 8 bins. Extraction time of LDGP is approximately $1/4^{th}$ of the extraction time of LVP and LDGP. Match time shown in Fig.6(a) is the match time for $2^{nd}$ order features extracted from $4 \times 4$ sub-regions with histogram of 8 bins. LDGP again outperforms LVP and LDP with respect to match time and extraction time.

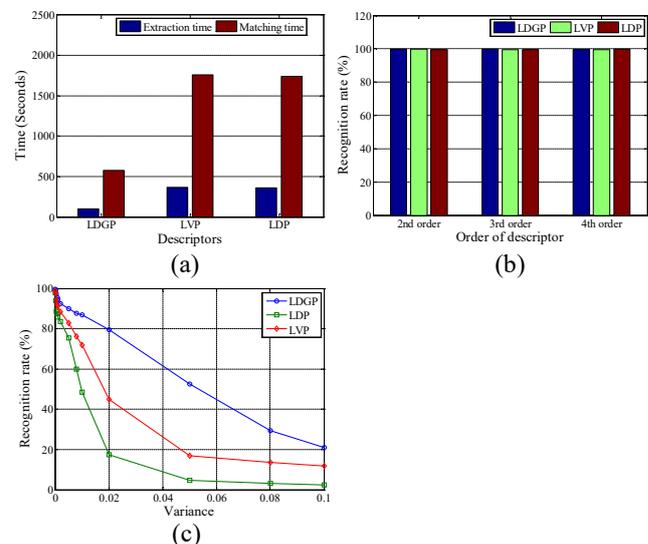

Fig. 6. Comparative recognition rates of LDGP, LVP and LDP on Extended Yale B Face datasets (a) with different order derivative space, (b) Extraction time and Match time of $2^{nd}$ order LDGP, LVP and LDP with $4 \times 4$ sub-regions and 8 histogram bins, (c) Comparative recognition rates of LDGP, LVP and LDP with different values of variance.

$2^{nd}$ order, $3^{rd}$ order and $4^{th}$ order features are extracted using LDGP, LVP and LDP in $1^{st}$ order, $2^{nd}$ order and $3^{rd}$ order derivative space respectively. The size of sub-regions and histogram are fixed to $4 \times 4$ and 8 bins respectively. Recognition rates for higher order features are shown in Fig.6(b). As shown in Fig.6(b) $2^{nd}$ and $3^{rd}$ order LDGP outperforms LDP and LVP with respect to recognition rates as well as extraction time and match time. 2nd and 3rd order descriptors perform better than 4th order descriptors as 4th order descriptors are more sensitive to noise [12].

The effect of increasing level of noise with varying illumination on the recognition rates of LDGP, LVP and LDP is shown in Fig.6(c). The percentage reduction in recognition accuracies of LDGP, LVP and LDP are 77.81%, 86.92% and 97.05% respectively with 99% increase in variance of Gaussian noise. Fig.6(c) shows that LDGP performs better than LVP and LDP under high level of noise with varying illuminations.

## C. Performance analysis on CMU-PIE database

The database contains image of 68 subjects taken under varying pose, illuminations, expression and lighting. Expression dataset contains 39 images with different expressions and pose for each subject. Illumination set of the database contains images with 24 different illuminations on 13 different pose of each individual. 72 images in 24 different lighting conditions with 3 different pose are in the lights dataset of the database. Talking dataset contains images with 60 different expressions and 3 different poses.

Probe sets are prepared by randomly selecting 20%, 30%, 40%, 50% and 60% images from each datasets and the remaining images are used as corresponding gallery for the respective datasets. 10 fold cross validation is used to calculate the average recognition rates for each probe and gallery pair of all the datasets. Average recognition rate is the average of recognition rates obtained in 10 iterations of the experiment for a particular probe and gallery pair. Fig.7(a-d) show the average recognition rates computed for different probe and gallery pairs of all the datasets. Average recognition rates of all three descriptors are 100% on lights and taking datasets as shown in Fig.7(b) and Fig.7(d) respectively. Average recognition rates of LDGP, LVP and LDP are comparable and almost consistent with varying size of probe and gallery for illumination dataset as shown in Fig.7(c). Average recognition rates of LDGP, LVP and LDP for expression dataset shown in Fig.7(a) are comparable.

Average match time to compute the average recognition rates on different datasets are shown in Fig.7(e-h). LDGP outperforms LVP and LDP with respect to match time for all the datasets. Mach time of LDGP is approximately 1/4th of the match times of LVP as well as LDP. LDGP achieves significant reduction in match time for large databases and it also maintains the accuracy of recognition.

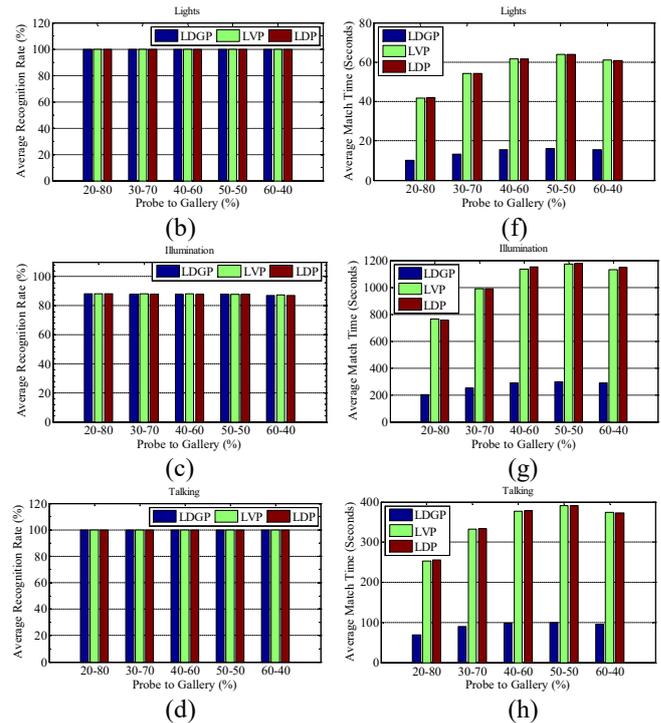

Fig. 7. Comparative average recognition rates of LDGP, LVP and LDP of 10-fold cross validation with different size probe and gallery on CMU_PIE datasets (a) different Expressions, (b) different lighting, (c) different illuminations, (d) talking. Comparative match time of LDGP, LVP and LDP of 10-fold cross validation with different size probe and gallery of CMU_PIE datasets (e) Expression (f) lights (g) illuminations, (h) talking.

## VI. CONCLUSION

In this paper a feature descriptor is proposed in the higher order derivative space. Proposed descriptor effectively eliminates problem of feature length increasing. Efficiency of the proposed descriptor is tested with respect to match time and extraction time. Proposed descriptor shows significant improvements in match time and extraction time. Its effectiveness and robustness are also tested against pose, illumination, expression and lighting variations in face recognition. Proposed descriptor achieves good recognition rates against pose, illumination, expression and lighting variations equivalent to the recognition accuracies achieved by the state of the art feature descriptors [11][12]. Finally the proposed descriptor is tested under severe noisy conditions. LDGP outperforms LDP and LVP under higher levels of noise.

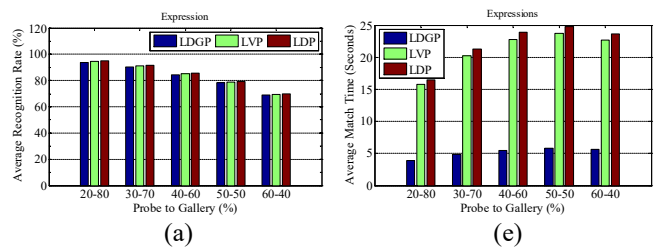

(a)                    (e)